\documentclass[letterpaper, 10 pt, conference]{ieeeconf}  

\IEEEoverridecommandlockouts                              

\usepackage{graphicx}
\usepackage{amsmath} 
\usepackage{amssymb}  

\usepackage{color}

\usepackage{multirow}

\newcommand\tabrotate[1]{\rotatebox{90}{#1\hspace{\tabcolsep}}}

\usepackage{fancyhdr}
\title{\LARGE \bf
Unsupervised Learning Methods for Visual Place Recognition \\ in Discretely and Continuously Changing Environments
}

\author{Stefan Schubert, Peer Neubert and Peter Protzel
\thanks{All authors are with Faculty of Electrical Engineering and Automation Technology, Chemnitz University of Technology, Chemnitz, Germany
        {\tt\small \{firstname.lastname\}@etit.tu-chemnitz.de}}%
}

\begin{document}

\maketitle
\thispagestyle{fancy}
\fancyhead[OL]{ 
	\footnotesize
	To appear in Proc. of IEEE International Conference on Robotics and Automation (ICRA), 2020, Paris, France. SUBMITTED VERSION\\
	\tiny
	\copyright 2020 IEEE. Personal use of this material is permitted. Permission from IEEE must be obtained for all other uses, in any current or future media, including
	reprinting/republishing this material for advertising or promotional purposes, creating new collective works, for resale or redistribution to servers or lists, or reuse of any copyrighted component of this work in other works.}

\addtolength{\headheight}{\baselineskip}

\begin{abstract}
Visual place recognition in changing environments is the problem of finding matchings between two sets of observations, a query set and a reference set, despite severe appearance changes.
Recently, image comparison using CNN-based descriptors showed very promising results.
However, the experiments in the literature typically assume a \textit{single} distinctive condition within each set (e.g., reference images are captured at daytime and the query sequence is at night).
In this paper, we will demonstrate that as soon as the conditions \textit{change within one set} (e.g., reference is daytime and now the query is a traversal daytime-dusk-night-dawn), different places under the same condition can suddenly look more similar than same places under different conditions.
As a consequence, state-of-the-art approaches like CNN-based descriptors fail.
This paper discusses this practically very important problem of \textit{in-sequence condition changes}
and defines a hierarchy of problem setups from (1) no in-sequence changes, (2) discrete in-sequence changes, to (3) continuous in-sequence changes.
We will experimentally evaluate the effect of in-sequence condition changes on two state-of-the-art CNN-descriptors and investigate unsupervised methods to improve their performance.
This includes an evaluation of the importance of statistical normalization (standardization) of descriptors, which is often omitted in existing approaches but can considerably improve results for problems up to discrete in-sequence changes.
To address the practical most relevant setup of continuous changes, we investigate the application of unsupervised learning methods using two PCA-based approaches from the literature and propose a novel clustering-based extension of the statistical normalization. 
We experimentally demonstrate that these approaches can significantly improve place recognition performance in case of continuous in-sequence condition changes.
With publication of this paper, we will provide the datasets and implementations.
\end{abstract}

\section{INTRODUCTION} \label{sec:intro}

Visual place recognition in changing environments is the challenging problem of recognizing places despite severe appearance changes, e.g., due to changing illumination, time of day, weather or season. 
Since it is an important mean for simultaneous localization and mapping (SLAM) it is an active field of research.
A barely described and addressed problem in existing research on visual place recognition in changing environments are in-sequence condition changes within the query- and/or reference-set (e.g., query: dusk-night-dawn, reference: day).
A potential consequence of unaddressed in-sequence condition changes was described by Vysotska and Stachniss in their recent RA-L paper \cite{Vysotska2019}:

\begingroup
\leftskip=0.2cm
\noindent ``\textit{[...] we observe a performance degradation whenever the visual appearance changes within the query sequence. For example, if the sequence starts at the evening and matching continues for a long time, so that it gets dark outside, the same non-matching parameter that reasonably described the non-matchiness of the sequence is no longer valid.}''
\endgroup

\begin{figure*}[tb]
 \centering
 \includegraphics[width=0.8\linewidth]{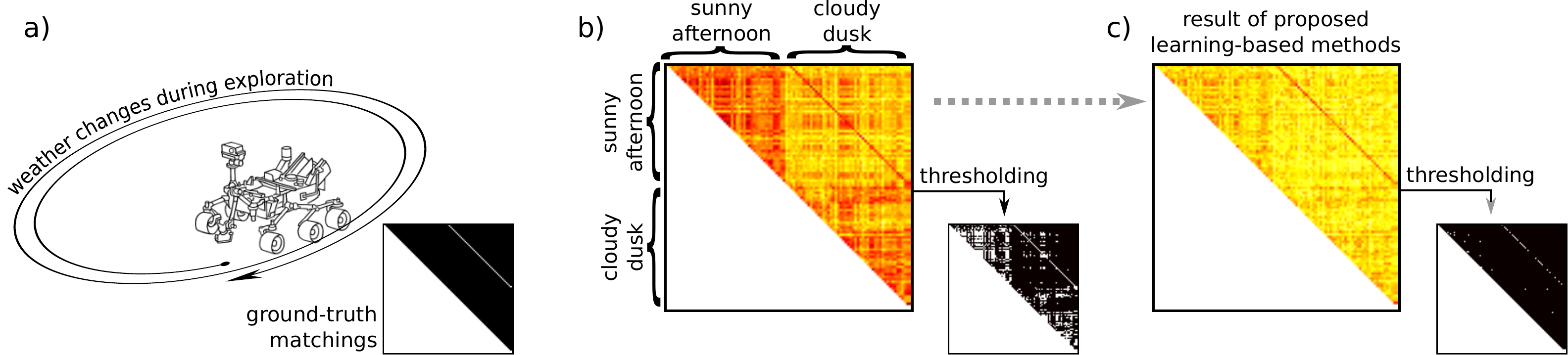}
 
 \vspace{-0.3cm}

 \caption{A robot explores an unknown environment while the environmental conditions change. The secondary diagonal in the ground truth represents the second loop matched to the first loop in the example. The image similarity matrix in b) illustrates the main problem: images of different places taken under the same environmental conditions are more similar than images of the same place under different environmental conditions. Although the true matchings on the secondary diagonal have a considerably high similarity (red color) compared to other comparisons between \textit{different} conditions, there are various comparable similar wrong matchings between the \textit{same} conditions. c) An unsupervised learning based method from this paper reduces the similarity of different places and improves the place matchings after thresholding.}

 \vspace{-0.65cm}
 
 \label{fig:intro}
\end{figure*}

Although they are only very rarely addressed in experimental evaluations, presumably, such in-sequence condition changes constitute a problem for many existing approaches.
The underlying problem is depicted in Fig.~\ref{fig:intro}: When the condition changes within a sequence and images are still compared with techniques that are not suited for varying conditions within a sequence, different places under the same condition can appear more similar than the same places under different conditions -- the visual place recognition performance drops tremendously.

In this paper, we define a hierarchy of in-sequence condition changes in Sec.~\ref{sec:cond_change} and give an overview of related work in Sec.~\ref{sec:related_work}.
In Sec.~\ref{sec:approaches} we propose a novel and describe two existing methods from the literature that are suited for continuous in-sequence condition changes.  
In particular, they do not rely on prior knowledge about occurring conditions.
Instead, they exploit unsupervised learning techniques that discover the effect of different conditions in a sequence and create more condition-invariant descriptors.
The algorithms do not build upon additional information like video sequences (i.e. adjacent frames are spatially adjacent in the real world) nor require in-sequence loops (i.e. frequent revisits of same places). However, the proposed methods can be combined with such approaches.
In our experiments in Sec.~\ref{sec:experiments}, we first evaluate the place recognition performance of two state-of-the-art CNN-descriptors on different datasets without in-sequence condition changes.
Here, we emphasize the importance of descriptor standardization which is often omitted in existing approaches.
However, in case of in-sequence condition changes the performance of raw and standardized CNN-descriptor degrades. 
Finally, we evaluate the three algorithms from Sec.~\ref{sec:approaches} on datasets with multiple in-sequence condition changes, and demonstrate that they can achieve remarkable improvements over the raw and standardized CNN-descriptors.

\section{DISCRETELY AND CONTINUOUSLY CHANGING ENVIRONMENTS} \label{sec:cond_change}
For visual place recognition in changing environments, we can distinguish three different types of in-sequences condition changes:
\begin{enumerate}
 \item \textit{Constant / in-sequence distinct condition:} Each, the query- and the reference-set, was recorded under one specific condition (e.g., query: winter, reference: summer).
 There is no in-sequence condition change.
 \item \textit{Discretely:} There are in-sequence condition changes in the query- and/or reference-set (e.g., query: summer-fall-winter-spring, reference: summer).
 The single conditions are partially constant.
 It is known which parts of the databases belong to one specific condition. 
 \item \textit{Continuously:} There are in-sequence condition changes in the query- and/or reference-set (e.g., query: summer-fall-winter-spring, reference: summer).
 The single conditions need not to be partially constant -- a condition change can happen continuously with ``intermediate conditions'' between two conditions.
 In general, the number of appearing conditions as well as the mapping between parts of the set and conditions is unknown.
\end{enumerate}

Continuous condition changes (3) can be considered as the most general case that includes discrete condition changes (2) and no condition changes (1) within each set.
Potentially, case (1) and (2) are merely simplifications of the real world and might not exist in uncontrolled outdoor environments.
Especially the discrete classification of conditions is probably hard and inaccurate: E.g., should we distinguish between lighting conditions at 1pm and 2pm; can we classify soft and heavy rain as rain; are both blue sky and cloudy sky sunny?

\vspace{-0.3cm}
\section{RELATED WORK}\label{sec:related_work}
\vspace{-0.1cm}
Visual place recognition in changing environments is a subject of active research while in-sequence condition changes achieved only little attention so far.
An overview of existing methods is given in \cite{Lowry2016a}.

Place descriptors can be used to describe and match places.
In \cite{Suenderhauf2015} it is shown that intermediate CNN-layer outputs like the \textit{conv3}-layer from AlexNet \cite{Krizhevsky2012} trained for image classification can serve as holistic image descriptors to match places across condition changes between reference and query.
Moreover, there are CNNs especially trained for place recognition that return either holistic image descriptors like NetVLAD \cite{Arandjelovic2018} or local features like DELF (DEep Local Features) \cite{Noh2017}.
Garg et al. \cite{Garg2018} use feature standardization to further improve the results that can be obtained with holistic CNN-descriptors.

Building upon such image descriptors, unsupervised learning techniques based on PCA (Principal Component Analysis) have been used.
In \cite{Liu2012}, Liu and Zhang show that a PCA-based dimensionality reduction on a Gabor-Gist descriptor from 960 dimensions to the 60 most discriminative elements (with the highest eigenvalues) can be used to perform loop closure detection computationally and memory efficient.
Further, they give a clue that descriptor comparisons between places for loop closure detection become more discriminative.
Lowry and Milford \cite{Lowry2016} proposed \textit{change removal} which is a similar dimensionality reduction approach.
However, instead of removing the less discriminative elements of a descriptor they show that removing the most discriminative elements leads to an improved place recognition performance for changing conditions between query and reference
The most discriminative principal components are supposed to contain place independent effects on the images caused by the conditions.
They also applied whitening to the new descriptors in PCA-space, which resulted in a performance degradation.

Instead of addressing the descriptors for condition-invariance, the images can be directly modified to emulate equal conditions between query and reference.
Milford et al. \cite{Milford2015} proposed the usage of a deep convolutional neural field model to estimate for all RGB-image the corresponding depth images that are potentially condition-invariant.
A distinct technique to achieve condition-invariance between images is \textit{appearance change prediction} that transfers images between predefined conditions.
Using pairs of images of same places under different conditions, Neubert et al. \cite{Neubert2013} proposed to learn a mapping of visual words between two conditions.
Similarly, \cite{Lowry2016} proposed to build a linear regression model with image pairs of the same place under two conditions.
In \cite{Anoosheh2018, Anoosheh2019}, Anoosheh et al. use a CycleGAN to train a generator that transfers images between predefined conditions.
The advantage of using CycleGANs \cite{Zhu2017} is that only a weak supervision is required -- instead of image pairs of the same place as training data, the system requires only sets of images that belong to a same meta-class like ``day'' and ``night''.

To address the problem of condition changes within sequences, Churchill et al. \cite{Churchill2013} proposed \textit{experience maps}.
Experience maps can cope with in-sequence condition changes but rely on loops and frequent revisits of places in the data as they need to observe places under slightly changing conditions to accumulate condition-variant descriptors for a condition-invariant representation.
Similarly, co-occurrence maps \cite{Johns2013} can be build that represent each place within one matrix with a bag of words approach and the feature's relative positions; again the representation accumulates observations of the same place during revisits with slight condition change.
McManus et al. \cite{McManus2014} learn offline condition-invariant broad-region detectors from beforehand collected images with a variety of appearances at particular locations.
Vysotska and Stachniss \cite{Vysotska2019} presented Multi-Sequence Maps that exploit discretely changing conditions within the reference.
The reference consists of multiple trajectories in an environment, each under the same condition.
Place matchings are performed with a graph search.
However, as quoted in Sec.~\ref{sec:intro} their system's performance could suffer from continuous condition changes within one trajectory.

Another technique to tackle a specific type of continuous condition changes are illumination invariant image conversion and shadow removal.
Illumination invariant image conversion \cite{Alvarez2011, Ranganathan2013, Maddern2014} transfers images into an illumination-invariant representation.
However, \cite{Maddern2014} shows that the assumption of a black-body illumination is violated, e.g., at night, and that therefore the illumination invariant images yield poor results.
In addition to illumination invariance, shadow removal \cite{Corke2013, Shakeri2016, Ying2016} can be used to get images that are illumination invariant and independent of sun positions.

\section{UNSUPERVISED LEARNING METHODS FOR CONTINUOUSLY CHANGING ENVIRONMENTS} \label{sec:approaches}

In this section, we describe one novel algorithm based on clustering and two PCA-based algorithms from the literature that are suited for place recognition in continuously changing environments.
This choice is based on the following restrictions and requirements:
\begin{enumerate}
  \item As a consequence of a continuously changing environment, we do not know the occurring conditions and the way they change. 
  Therefore, appearance change predictions with condition classification and transfer models like \cite{Neubert2013, Anoosheh2019} cannot be applied as these may not be available for all potentially occurring conditions. Instead, unsupervised learning techniques have to be used to discover structures in the data caused by arbitrary conditions.
  \item Condition changes beyond illumination changes could occur, so that illumination invariant image conversion and shadow removal may not be sufficient.
  \item We solely use ``image-only information''. The algorithms must not use additional sensor information or sequence information.
  Loops / multiple revisits need not necessarily occur.
  \item Image descriptors have to be converted into more condition-independent descriptors (no distinct representation).
  Therefore, sequence-based methods or experience maps could be used as a subsequent step to improve results.
\end{enumerate}

To match all requirements, we choose unsupervised learning based methods as they are potentially suited for finding structures in the data caused by different conditions including ``intermediate conditions''.
Every algorithm below performs a mapping from a less condition-invariant set of descriptors $D$ to a potentially more condition-invariant descriptor set $D'$.

\subsection{Standardization of Descriptors (STD) and K Standardizations of Descriptors (K-STD)} \label{sec:k}
CNN-descriptors showed impressing performance for place recognition in changing environments, e.g. \cite{Suenderhauf2015}.
Statistical standardization of descriptors (STD) is a simple technique that can further significantly improve the results - however, it is often omitted in existing approaches.
We apply this technique to the CNN-descriptors and use it as a baseline which achieves state-of-the-art performance for place recognition tasks in changing environments without in-sequence condition changes.

STD is the standardization of the descriptors $D_i$ in the set $D$ of all descriptors.
The standardization is performed separately on query and reference with the following equation:
\vspace{-0.5cm}
\begin{align}\label{eq:NSD}
  D_i' = \frac{D_i - \mu_S}{\sigma_S} \hspace{1cm} \forall i = 1 \ldots |D|
\end{align}
Here, $\mu_S$ and $\sigma_S$ are the mean and the standard deviation of $D$, respectively, across features.

\begin{figure}[tb]
 \centering
 \includegraphics[width=\linewidth]{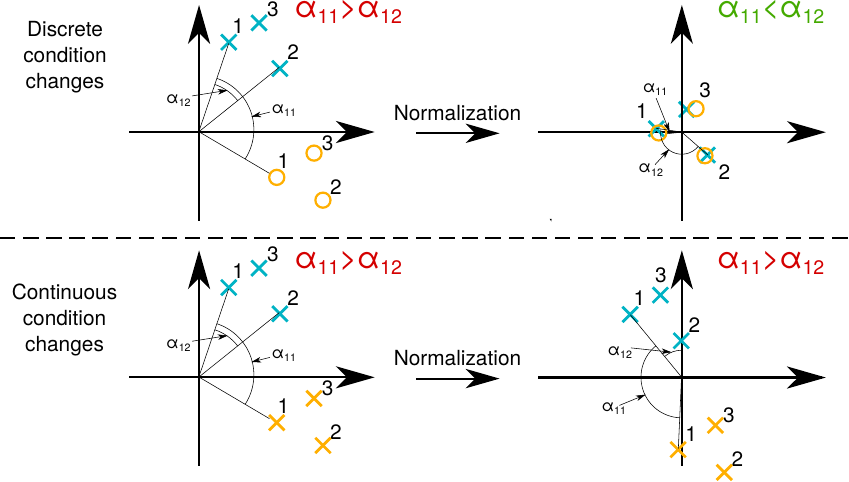}
 \vspace{-0.8cm}
 \caption{Illustration of the problem with standardization for discretely and continuously changing environments. Before normalization, the angle $\alpha$ (i.e. cosine distance) is lower between place 1 and 2 ($\alpha_{12}$) than between place 1 under different conditions ($\alpha_{11}$). After normalization, for discrete condition changes same places get more similar and different places more different, but for continuous condition changes the actual similarities cannot be resolved. By finding clusters, the same performance as for discrete condition changes could be recovered. Crosses and circles illustrate knowledge about different conditions; in the continuous case, we do not have this knowledge.}
 \vspace{-0.5cm}
 \label{fig:idea_KNSD}
\end{figure}

However, the results in Sec.~\ref{sec:raw_descriptors} will show that the standardization fails under continuously changing conditions.
Therefore, we propose an unsupervised learning approach that is based on standardization called \textit{K Standardizations of Descriptors} (K-STD).
The underlying idea is illustrated in Fig.~\ref{fig:idea_KNSD}: The descriptors of the same places under different conditions could have a systematic condition-dependent offset in feature space.
K-STD tries to find $K$ clusters in $D$ to perform a standardization on each of these subsets $D^k$ separately:
\vspace{-0.4cm}
\begin{align}
  D^k = \text{cluster}(D, K)
\end{align}
A suitably large $K$ has to be used to address continuous changes with a discrete set of $K$ clusters.
In our experiments, we use k-means (with cosine distance) to find $K$ sets of descriptors $D^k$ in $D$.
Subsequently, each cluster is standardized like in Eq.~\ref{eq:NSD}, and combined to the more condition-invariant set of descriptors $D'$:
\vspace{-0.1cm}
\begin{gather}
  D_i^{\prime k} = \frac{D^k_i - \mu^k_S}{\sigma^k_S} \hspace{1cm} \forall i = 1 \ldots |D^k|, \hspace{0.2cm} \forall k = 1 \ldots K \\
  D' = \{D^{\prime k} \hspace{0.15cm} | \hspace{0.15cm} \forall k = 1 \ldots K\}
\end{gather}
$\mu^k_S$ and $\sigma^k_S$ are again the mean and the standard deviation of $D^k$, respectively.

\subsection{Dimensionality Reduction (DR) \& Change Removal (CR)}
As described in Sec.~\ref{sec:related_work}, both \textit{Dimensionality Reduction} (DR) \cite{Liu2012} and \textit{Change Removal} (CR) \cite{Lowry2016} are techniques that use a Principal Component Analysis (PCA) to improve the quality of place matchings.
In contrast to STD and K-STD, these approaches assume a set of descriptors $D$ that is the union of query and reference.

Both DR and CR start with a PCA to transform each descriptors $D_i$ into the principal component space.
The principal components $V$ can be calculated from a Singular Value Decomposition (SVD).
Note that $D$ should be standardized in advance.
\vspace{-0.3cm}
\begin{gather}
 [U,S,V] = \text{SVD}(D) \\
 D_i'' = D_i^T \cdot V
\hspace{0.3cm}\text{ with }
  D_i'' = (d^{\prime\prime i}_1, \ldots, d^{\prime\prime i}_{|D_i''|})^T
\end{gather}
The lower the index $j$ for the coefficient $d^{\prime\prime i}_j$ is, the higher is the descriptive meaning (i.e., the higher the eigenvalue).

For DR, the $q$ most descriptive principal components are kept:
\vspace{-0.3cm}
\begin{align}
 D_i' = (d^{\prime\prime i}_1, \ldots, d^{\prime\prime i}_{q})^T \hspace{0.3cm}\text{ with } q \leq |D_i''|
\end{align}

For CR, the $p$ most descriptive principal components are removed:
\vspace{-0.3cm}
\begin{align}
 D_i' = (d^{\prime\prime i}_{p+1}, \ldots, d^{\prime\prime i}_{|D_i''|})^T \hspace{0.3cm}\text{ with } p \geq 0
\end{align}

To use DR and CR jointly, $q$ and $p$ can be combined:
\vspace{-0.1cm}
\begin{align}
 D_i' = (d^{\prime\prime i}_{p+1}, \ldots, d^{\prime\prime i}_{q})^T
\end{align}

Finally, whitening could be used to give all coefficients the same variance
\vspace{-0.3cm}
\begin{align}
 D_i' := \frac{D_i'}{\sigma_S}
\end{align}
where $\sigma_S$ is the standard deviation of $D'$.

\section{EXPERIMENTAL RESULTS} \label{sec:experiments}

\definecolor{supergood}{rgb}{.2,.6,.2} 
\definecolor{good}{rgb}{.5,.99,.5} 
\definecolor{equal}{rgb}{0.1,0.1,0.1}  
\definecolor{bad}{rgb}{.9,.7,.1} 
\definecolor{superbad}{rgb}{.9,.1,.1}   
\definecolor{notworking}{rgb}{.5,.5,.5}
\fboxrule0.65pt

In this section, we evaluate raw and standardized CNN-descriptors on datasets without and with in-sequence condition changes, and investigate the performance of the three unsupervised learning based methods from Sec.~\ref{sec:approaches} on datasets with continuously changing conditions.

\vspace{-0.1cm}
\subsection{Experimental Setup}
NetVLAD \cite{Arandjelovic2018} and AlexNet \cite{Krizhevsky2012} are used as CNN-descriptors in the following experiments. For NetVLAD, we use the author's implementation trained on the Pitts30k dataset with VGG-16 and whitening that returns a $4096$-dimensional descriptor. For AlexNet, we use the conv3-layer output of Matlab's ImageNet model -- the resulting $64896$-dimensional descriptor is projected onto a $8192$-dimensional vector with Locality Sensitive Hashing (LSH); LSH is a random projection with normalized vectors drawn from a normal distribution.
We use \textit{average precision} (avgP) as performance metric for all experiments.
The \textit{cosine similarity} is used to measure the similarity between two descriptors.

\subsection{Datasets}

All evaluations are based on six different datasets with different characteristics regarding the environment, appearance changes, in-sequence loops, stops, or viewpoint changes.
\textbf{GardensPoint} Walking \cite{Glover2014}: Recorded with a hand-held camera on a single route on campus, twice during daytime on the left and right side and once at night on the right side.
\textbf{StLucia} (Various Times of the Day) \cite{Glover2010}: Collected with a webcam mounted on a car driving in a suburb from morning to afternoon over several days.
\textbf{CMU} Visual Localization \cite{Badino2011}: Five car rides along a 8km route with stops, weather, and seasonal changes. We use the left camera.
\textbf{Oxford} RobotCar \cite{Maddern2017}: Car rides collected over one year under several conditions. We use the center camera of the trinocular stereo camera.
\textbf{Nordland} \cite{Suenderhauf2013}: Viewpoint aligned rides along a train track once per season. We use a subset of $288$ distinct places.
\textbf{SFU} Mountain \cite{Bruce2015}: Multiple traverses of a mobile robot in semi-structured woodland under varying lighting and weather conditions.

Based on these datasets, we define three dataset combinations with multiple sequence combinations: 
1) \textbf{Single-condition datasets} -- Each set, reference and query, contains only one specific condition -- this is the case for most place recognition setups in the literature.
2) \textbf{Two-conditions datasets} -- The datasets contain the same sequence-combinations as the single-condition datasets, but reference and query are concatenated, i.e. both are identical.
3) \textbf{Multi-conditions datasets} -- Reference and query are identical and contain multiple conditions.
We define three multi-conditions datasets:
Nordland due to the seasonal changes: \{spring $\rightarrow$ summer $\rightarrow$ fall $\rightarrow$ winter\} ($1152$ descriptors).
SFU because of its most unstructured environment: \{dry $\rightarrow$ dusk $\rightarrow$ january $\rightarrow$ november $\rightarrow$ september $\rightarrow$ wet\} ($2310$ descriptors).
Oxford as one of the car-captured datasets with changing conditions from sunny daytime to night: \{sun $\rightarrow$ clouds $\rightarrow$ overcast $\rightarrow$ rain $\rightarrow$ dusk $\rightarrow$ night\}\footnote{Oxford RobotCar sequences: sun (2014-12-16-09-14-09), clouds (2014-11-18-13-20-12), overcast (2014-12-09-13-21-02), rain (2014-12-05-11-09-10), dusk (2015-02-20-16-34-06), night (2014-12-16-18-44-24)} ($13352$ descriptors).
Dependent on whether the types of different conditions and the actual condition of a query image are known or not, the multi-condition datasets can represent discrete or continuous in-sequence condition changes.

\subsection{Raw and standardized CNN-descriptors under in-sequence condition changes} \label{sec:raw_descriptors}
First, we evaluate the performance of the raw and standardized CNN-descriptors without condition changes on the \textit{single-condition datasets}; results are shown in Table~\ref{tab:split} (top).
The results show the relatively good results of the descriptors and that the standardization of descriptors maintains or often improves the performance.
This emphasizes that standardization should always be used for place recognition without in-sequence condition changes which is often not the case in experiments from the literature.

Table~\ref{tab:split} (mid, bottom) shows results under discrete and continuous condition changes on the \textit{two-conditions datasets}.
For the discrete condition changes, we exploit knowledge about the conditions within the sequences: the descriptors are standardized separately for each condition.
For the continuous condition changes, the standardization cannot build upon this knowledge: the standardization is conducted once for all descriptors independent of their condition.
The results in Table~\ref{tab:split} reveal that the performance for the raw descriptors drops in case of in-sequence condition changes.
While the performance of the standardized descriptors under discrete condition changes decreases only slightly, the performance under continuous condition changes drops similar to the raw descriptors.
Thus, alternative methods are required in case of continuous condition changes to achieve reasonable results.

\boldmath
\begin{table}[tb]
\centering
\caption{Evaluation of the average precision on the single-condition and two-conditions datasets for the three types of in-sequence condition changes.}
\vspace{-0.1cm}
\resizebox{0.45\textwidth}{!}{%
\tabrotate{in-sequence distinct condition}
  \begin{tabular}{llllllll}
  \hline
                   &             &                 & \multicolumn{2}{c}{\textbf{NetVLAD}}             & \multicolumn{2}{c}{\textbf{AlexNet - LSH}} \\
\textbf{Dataset} & \textbf{Reference} & \textbf{Query}  & \textbf{Raw}  & \textbf{STD}  & \textbf{Raw}  & \textbf{STD} \\
\hline       
Nordland & fall & spring & 0.39 \color{equal}{} &  0.61 \color{supergood}{$\uparrow$} &  0.82 \color{equal}{} &  \textbf{0.92} \color{good}{$\nearrow$}  \\
 & fall & winter & 0.06 \color{equal}{} &  0.26 \color{supergood}{$\uparrow$} &  0.63 \color{equal}{} &  \textbf{0.83} \color{supergood}{$\uparrow$}  \\
 & spring & winter & 0.11 \color{equal}{} &  0.37 \color{supergood}{$\uparrow$} &  0.59 \color{equal}{} &  \textbf{0.86} \color{supergood}{$\uparrow$}  \\
 & summer & spring & 0.32 \color{equal}{} &  0.58 \color{supergood}{$\uparrow$} &  0.77 \color{equal}{} &  \textbf{0.90} \color{good}{$\nearrow$}  \\
 & summer & fall & 0.63 \color{equal}{} &  0.84 \color{supergood}{$\uparrow$} &  0.94 \color{equal}{} &  \textbf{0.97} \color{equal}{$\rightarrow$}  \\
StLucia & 100909-0845 & 190809-0845 & 0.41 \color{equal}{} &  0.58 \color{supergood}{$\uparrow$} &  0.59 \color{equal}{} &  \textbf{0.66} \color{good}{$\nearrow$}  \\
 & 100909-1000 & 210809-1000 & 0.47 \color{equal}{} &  0.61 \color{supergood}{$\uparrow$} &  0.57 \color{equal}{} &  \textbf{0.67} \color{good}{$\nearrow$}  \\
 & 100909-1210 & 210809-1210 & 0.51 \color{equal}{} &  0.63 \color{good}{$\nearrow$} &  0.55 \color{equal}{} &  \textbf{0.67} \color{good}{$\nearrow$}  \\
 & 100909-1410 & 190809-1410 & 0.38 \color{equal}{} &  0.57 \color{supergood}{$\uparrow$} &  0.61 \color{equal}{} &  \textbf{0.70} \color{good}{$\nearrow$}  \\
 & 110909-1545 & 180809-1545 & 0.27 \color{equal}{} &  0.44 \color{supergood}{$\uparrow$} &  0.60 \color{equal}{} &  \textbf{0.67} \color{good}{$\nearrow$}  \\
CMU & 20110421 & 20100901 & \textbf{0.73} \color{equal}{} &  \textbf{0.73} \color{equal}{$\rightarrow$} &  0.43 \color{equal}{} &  0.59 \color{supergood}{$\uparrow$}  \\
 & 20110421 & 20100915 & 0.77 \color{equal}{} &  \textbf{0.78} \color{equal}{$\rightarrow$} &  0.58 \color{equal}{} &  0.66 \color{good}{$\nearrow$}  \\
 & 20110421 & 20101221 & 0.56 \color{equal}{} &  \textbf{0.59} \color{equal}{$\rightarrow$} &  0.33 \color{equal}{} &  0.48 \color{supergood}{$\uparrow$}  \\
 & 20110421 & 20110202 & 0.61 \color{equal}{} &  \textbf{0.71} \color{good}{$\nearrow$} &  0.32 \color{equal}{} &  0.47 \color{supergood}{$\uparrow$}  \\
GardensPoint & day-right & day-left & 0.97 \color{equal}{} &  \textbf{0.99} \color{equal}{$\rightarrow$} &  0.58 \color{equal}{} &  0.60 \color{equal}{$\rightarrow$}  \\
Walking & day-right & night-right & 0.51 \color{equal}{} &  0.68 \color{supergood}{$\uparrow$} &  0.51 \color{equal}{} &  \textbf{0.69} \color{supergood}{$\uparrow$}  \\
 & day-left & night-right & 0.40 \color{equal}{} &  \textbf{0.56} \color{supergood}{$\uparrow$} &  0.10 \color{equal}{} &  0.25 \color{supergood}{$\uparrow$}  \\
Oxford & 141209 & 141216 & 0.87 \color{equal}{} &  \textbf{0.92} \color{equal}{$\rightarrow$} &  0.49 \color{equal}{} &  0.65 \color{supergood}{$\uparrow$}  \\
 & 141209 & 150203 & 0.93 \color{equal}{} &  \textbf{0.96} \color{equal}{$\rightarrow$} &  0.63 \color{equal}{} &  0.85 \color{supergood}{$\uparrow$}  \\
 & 141209 & 150519 & 0.83 \color{equal}{} &  \textbf{0.91} \color{equal}{$\rightarrow$} &  0.25 \color{equal}{} &  0.80 \color{supergood}{$\uparrow$}  \\
 & 150519 & 150203 & 0.85 \color{equal}{} &  \textbf{0.94} \color{good}{$\nearrow$} &  0.30 \color{equal}{} &  0.89 \color{supergood}{$\uparrow$}  \\

  \hline
  \end{tabular}
}

\resizebox{0.45\textwidth}{!}{%
\tabrotate{discrete condition change}
  \begin{tabular}{llllllll}
  \hline
                   &             &                 & \multicolumn{2}{c}{\textbf{NetVLAD}}             & \multicolumn{2}{c}{\textbf{AlexNet - LSH}} \\
\textbf{Dataset} & \multicolumn{2}{c}{\textbf{Reference/Query}} & \textbf{Raw}  & \textbf{STD}  & \textbf{Raw}  & \textbf{STD} \\
\hline       
Nordland & fall & spring & 0.13 \color{equal}{} &  0.41 \color{supergood}{$\uparrow$} &  0.49 \color{equal}{} &  \textbf{0.70} \color{supergood}{$\uparrow$}  \\
 & fall & winter & 0.00 \color{equal}{} &  0.03 \color{notworking}{$\times$} &  0.02 \color{equal}{} &  \textbf{0.33} \color{supergood}{$\uparrow$}  \\
 & spring & winter & 0.00 \color{equal}{} &  0.06 \color{notworking}{$\times$} &  0.05 \color{equal}{} &  \textbf{0.42} \color{supergood}{$\uparrow$}  \\
 & summer & spring & 0.06 \color{equal}{} &  0.35 \color{supergood}{$\uparrow$} &  0.38 \color{equal}{} &  \textbf{0.69} \color{supergood}{$\uparrow$}  \\
 & summer & fall & 0.53 \color{equal}{} &  0.77 \color{supergood}{$\uparrow$} &  0.89 \color{equal}{} &  \textbf{0.93} \color{equal}{$\rightarrow$}  \\
StLucia & 100909-0845 & 190809-0845 & 0.34 \color{equal}{} &  0.51 \color{supergood}{$\uparrow$} &  0.54 \color{equal}{} &  \textbf{0.60} \color{good}{$\nearrow$}  \\
 & 100909-1000 & 210809-1000 & 0.45 \color{equal}{} &  0.57 \color{supergood}{$\uparrow$} &  0.52 \color{equal}{} &  \textbf{0.62} \color{good}{$\nearrow$}  \\
 & 100909-1210 & 210809-1210 & 0.53 \color{equal}{} &  0.64 \color{good}{$\nearrow$} &  0.57 \color{equal}{} &  \textbf{0.68} \color{good}{$\nearrow$}  \\
 & 100909-1410 & 190809-1410 & 0.32 \color{equal}{} &  0.49 \color{supergood}{$\uparrow$} &  0.56 \color{equal}{} &  \textbf{0.65} \color{good}{$\nearrow$}  \\
 & 110909-1545 & 180809-1545 & 0.27 \color{equal}{} &  0.43 \color{supergood}{$\uparrow$} &  0.57 \color{equal}{} &  \textbf{0.63} \color{good}{$\nearrow$}  \\
CMU & 20110421 & 20100901 & 0.44 \color{equal}{} &  \textbf{0.50} \color{good}{$\nearrow$} &  0.21 \color{equal}{} &  0.35 \color{supergood}{$\uparrow$}  \\
 & 20110421 & 20100915 & 0.62 \color{equal}{} &  \textbf{0.63} \color{equal}{$\rightarrow$} &  0.40 \color{equal}{} &  0.48 \color{good}{$\nearrow$}  \\
 & 20110421 & 20101221 & 0.22 \color{equal}{} &  \textbf{0.33} \color{supergood}{$\uparrow$} &  0.10 \color{equal}{} &  0.18 \color{supergood}{$\uparrow$}  \\
 & 20110421 & 20110202 & 0.23 \color{equal}{} &  \textbf{0.35} \color{supergood}{$\uparrow$} &  0.08 \color{equal}{} &  0.21 \color{supergood}{$\uparrow$}  \\
GardensPoint & day-right & day-left & 0.90 \color{equal}{} &  \textbf{0.96} \color{equal}{$\rightarrow$} &  0.14 \color{equal}{} &  0.29 \color{supergood}{$\uparrow$}  \\
Walking & day-right & night-right & 0.01 \color{equal}{} &  0.23 \color{supergood}{$\uparrow$} &  0.22 \color{equal}{} &  \textbf{0.50} \color{supergood}{$\uparrow$}  \\
 & day-left & night-right & 0.01 \color{equal}{} &  \textbf{0.12} \color{supergood}{$\uparrow$} &  0.01 \color{equal}{} &  0.05 \color{notworking}{$\times$}  \\
Oxford & 141209 & 141216 & 0.68 \color{equal}{} &  \textbf{0.79} \color{good}{$\nearrow$} &  0.30 \color{equal}{} &  0.51 \color{supergood}{$\uparrow$}  \\
 & 141209 & 150203 & 0.75 \color{equal}{} &  \textbf{0.83} \color{good}{$\nearrow$} &  0.24 \color{equal}{} &  0.61 \color{supergood}{$\uparrow$}  \\
 & 141209 & 150519 & 0.66 \color{equal}{} &  \textbf{0.75} \color{good}{$\nearrow$} &  0.19 \color{equal}{} &  0.62 \color{supergood}{$\uparrow$}  \\
 & 150519 & 150203 & 0.27 \color{equal}{} &  \textbf{0.59} \color{supergood}{$\uparrow$} &  0.07 \color{equal}{} &  0.54 \color{supergood}{$\uparrow$}  \\

  \hline
  \end{tabular}
}

\resizebox{0.45\textwidth}{!}{%
\tabrotate{continuous condition change}
  \begin{tabular}{llllllll}
  \hline
                   &             &                 & \multicolumn{2}{c}{\textbf{NetVLAD}}             & \multicolumn{2}{c}{\textbf{AlexNet - LSH}} \\
\textbf{Dataset} & \multicolumn{2}{c}{\textbf{Reference/Query}}  & \textbf{Raw}  & \textbf{STD}  & \textbf{Raw}  & \textbf{STD} \\
\hline       
Nordland & fall & spring & 0.13 \color{equal}{} &  0.29 \color{supergood}{$\uparrow$} &  0.49 \color{equal}{} &  \textbf{0.54} \color{equal}{$\rightarrow$}  \\
 & fall & winter & 0.00 \color{equal}{} &  0.01 \color{notworking}{$\times$} &  0.02 \color{equal}{} &  \textbf{0.05} \color{notworking}{$\times$}  \\
 & spring & winter & 0.00 \color{equal}{} &  0.01 \color{notworking}{$\times$} &  0.05 \color{equal}{} &  \textbf{0.10} \color{notworking}{$\times$}  \\
 & summer & spring & 0.06 \color{equal}{} &  0.22 \color{supergood}{$\uparrow$} &  0.38 \color{equal}{} &  \textbf{0.49} \color{supergood}{$\uparrow$}  \\
 & summer & fall & 0.53 \color{equal}{} &  0.74 \color{supergood}{$\uparrow$} &  0.89 \color{equal}{} &  \textbf{0.92} \color{equal}{$\rightarrow$}  \\
StLucia & 100909-0845 & 190809-0845 & 0.34 \color{equal}{} &  0.47 \color{supergood}{$\uparrow$} &  0.54 \color{equal}{} &  \textbf{0.59} \color{good}{$\nearrow$}  \\
 & 100909-1000 & 210809-1000 & 0.45 \color{equal}{} &  0.56 \color{supergood}{$\uparrow$} &  0.52 \color{equal}{} &  \textbf{0.61} \color{good}{$\nearrow$}  \\
 & 100909-1210 & 210809-1210 & 0.53 \color{equal}{} &  0.63 \color{good}{$\nearrow$} &  0.57 \color{equal}{} &  \textbf{0.67} \color{good}{$\nearrow$}  \\
 & 100909-1410 & 190809-1410 & 0.32 \color{equal}{} &  0.45 \color{supergood}{$\uparrow$} &  0.56 \color{equal}{} &  \textbf{0.64} \color{good}{$\nearrow$}  \\
 & 110909-1545 & 180809-1545 & 0.27 \color{equal}{} &  0.41 \color{supergood}{$\uparrow$} &  0.57 \color{equal}{} &  \textbf{0.62} \color{equal}{$\rightarrow$}  \\
CMU & 20110421 & 20100901 & \textbf{0.44} \color{equal}{} &  0.41 \color{equal}{$\rightarrow$} &  0.21 \color{equal}{} &  0.26 \color{supergood}{$\uparrow$}  \\
 & 20110421 & 20100915 & \textbf{0.62} \color{equal}{} &  0.59 \color{equal}{$\rightarrow$} &  0.40 \color{equal}{} &  0.43 \color{equal}{$\rightarrow$}  \\
 & 20110421 & 20101221 & \textbf{0.22} \color{equal}{} &  0.17 \color{bad}{$\searrow$} &  0.10 \color{equal}{} &  0.09 \color{notworking}{$\times$}  \\
 & 20110421 & 20110202 & \textbf{0.23} \color{equal}{} &  \textbf{0.23} \color{equal}{$\rightarrow$} &  0.08 \color{equal}{} &  0.12 \color{supergood}{$\uparrow$}  \\
GardensPoint & day-right & day-left & 0.90 \color{equal}{} &  \textbf{0.95} \color{equal}{$\rightarrow$} &  0.14 \color{equal}{} &  0.20 \color{supergood}{$\uparrow$}  \\
Walking & day-right & night-right & 0.01 \color{equal}{} &  0.02 \color{notworking}{$\times$} &  0.22 \color{equal}{} &  \textbf{0.28} \color{supergood}{$\uparrow$}  \\
 & day-left & night-right & \textbf{0.01} \color{equal}{} &  \textbf{0.01} \color{notworking}{$\times$} &  \textbf{0.01} \color{equal}{} &  \textbf{0.01} \color{notworking}{$\times$}  \\
Oxford & 141209 & 141216 & \textbf{0.68} \color{equal}{} &  0.64 \color{equal}{$\rightarrow$} &  0.30 \color{equal}{} &  0.40 \color{supergood}{$\uparrow$}  \\
 & 141209 & 150203 & \textbf{0.75} \color{equal}{} &  0.72 \color{equal}{$\rightarrow$} &  0.24 \color{equal}{} &  0.37 \color{supergood}{$\uparrow$}  \\
 & 141209 & 150519 & \textbf{0.66} \color{equal}{} &  \textbf{0.66} \color{equal}{$\rightarrow$} &  0.19 \color{equal}{} &  0.48 \color{supergood}{$\uparrow$}  \\
 & 150519 & 150203 & 0.27 \color{equal}{} &  \textbf{0.30} \color{equal}{$\rightarrow$} &  0.07 \color{equal}{} &  0.25 \color{supergood}{$\uparrow$}  \\

  \hline
  \end{tabular}
}

\label{tab:split}
\end{table}
\unboldmath

\vspace{-0.1cm}
\subsection{Evaluation of approaches under continuous changes} \label{sec:exp_all}
The \textit{multi-conditions datasets} are used to investigate the performance of the three unsupervised learning based methods from Sec.~\ref{sec:approaches} together with the raw and standardized CNN-descriptors under continuous condition changes.
For the PCA-based approaches, we also investigate the influence of whitening that gives the descriptor elements unit variance in principal component space; this technique was also used in the experiments on CR in \cite{Lowry2016}.

Table~\ref{tab:max_performance} shows the maximum achievable results on the \textit{multi-conditions datasets} for all presented methods in the paper including the combination of DR and CR with optimal parameters $K$, $q$ or $p$.
The unsupervised learning based methods can improve the results remarkably in comparison to the raw and standardized descriptors.
None of the approaches performs always best, so that the used method has to be chosen carefully.
The proposed K-STD performs comparably to DR and CR.
The place recognition performance for DR was not evaluated in \cite{Liu2012}; the achieved results support that this approach is well suited for place recognition.
It is interesting that DR and CR perform equally, and that it is not important whether the most or least discriminative coefficients in principal component space are removed.
Table~\ref{tab:runtimes_memory} shows runtimes and memory consumptions for the three algorithms implemented in Matlab for different descriptor lengths and exemplary parameters.

\begin{table}[tb]
\caption{Average precision over algorithms with optimal parameters on the multi-conditions datasets with continuous condition changes.}
\vspace{-0.1cm}
\resizebox{0.5\textwidth}{!}{%
\begin{tabular}{lllllllllll}
\hline
                             & \textbf{}        & \textbf{}    & \textbf{}    & \textbf{}      & \multicolumn{3}{c}{\textbf{w/o whitening}}   & \multicolumn{3}{c}{\textbf{with whitening}}  \\
\textbf{Descriptor}          & \textbf{Dataset} & \textbf{Raw} & \textbf{STD} & \textbf{K-STD} & \textbf{DR} & \textbf{CR}   & \textbf{DR+CR} & \textbf{DR}   & \textbf{CR} & \textbf{DR+CR} \\ \hline
\multirow{3}{*}{AlexNet-LSH} & Nordland         & 0.37         & 0.46         & 0.74           & 0.45        & 0.80          & 0.81           & 0.83          & 0.00        & \textbf{0.84}  \\
                             & SFU              & 0.26         & 0.33         & 0.67           & 0.35        & 0.75          & 0.77           & \textbf{0.79} & 0.00        & \textbf{0.79}  \\
                             & Oxford           & 0.16         & 0.19         & 0.33           & 0.18        & \textbf{0.37} & \textbf{0.37}  & 0.23          & 0.03        & 0.26           \\ \hline
\multirow{3}{*}{NetVLAD}     & Nordland         & 0.14         & 0.22         & 0.37           & 0.23        & \textbf{0.41} & \textbf{0.41}  & 0.40          & 0.00        & \textbf{0.41}  \\
                             & SFU              & 0.12         & 0.18         & 0.35           & 0.19        & \textbf{0.38} & \textbf{0.38}  & 0.35          & 0.00        & 0.36           \\
                             & Oxford           & 0.23         & 0.21         & \textbf{0.51}  & 0.20        & 0.50          & 0.50           & 0.17          & 0.11        & 0.26          
\end{tabular}
}
\label{tab:max_performance}
\end{table}

\begin{table}[tb]
\caption{Runtime and memory consumption for finding clusters or principal components (``learning'') and for using it on new descriptors (``inference'') on the Oxford multi-conditions dataset. Implementations are in Matlab and ran on a Intel i7-5775C CPU.}
\vspace{-0.1cm}
\resizebox{0.5\textwidth}{!}{%
\begin{tabular}{l|ccc}
\hline
                   & \textbf{K-STD ($K = 20$)} & \textbf{DR ($q = 400$)} & \textbf{CR ($p = 15$)} \\ \hline
Descriptor Length  & \multicolumn{3}{c}{4096 / 8192 / 64896}                                      \\ \hline
Runtime Learning   & 5.4 / 11.1 / 94.4 sec     & 101 / 183 / 1380 sec    & 6.9 / 11 / 88.1 sec    \\
Runtime Inference  & 2.1 / 2.8 / 10.3 msec     & 0.9 / 1.9 / 13.8 msec   & 0.4 / 0.5 / 2.1 msec   \\
Memory Consumption & 1.3 / 2.5 / 19.8 MB       & 12.5 / 25.1 / 198.5 MB  & 0.5 / 1 / 7.9 MB      
\end{tabular}
}
\label{tab:runtimes_memory}
\end{table}

\subsection{Evaluation of parameter $K$ for K-STD}
\begin{figure}[tb]
 \centering

 \includegraphics[width=0.45\linewidth]{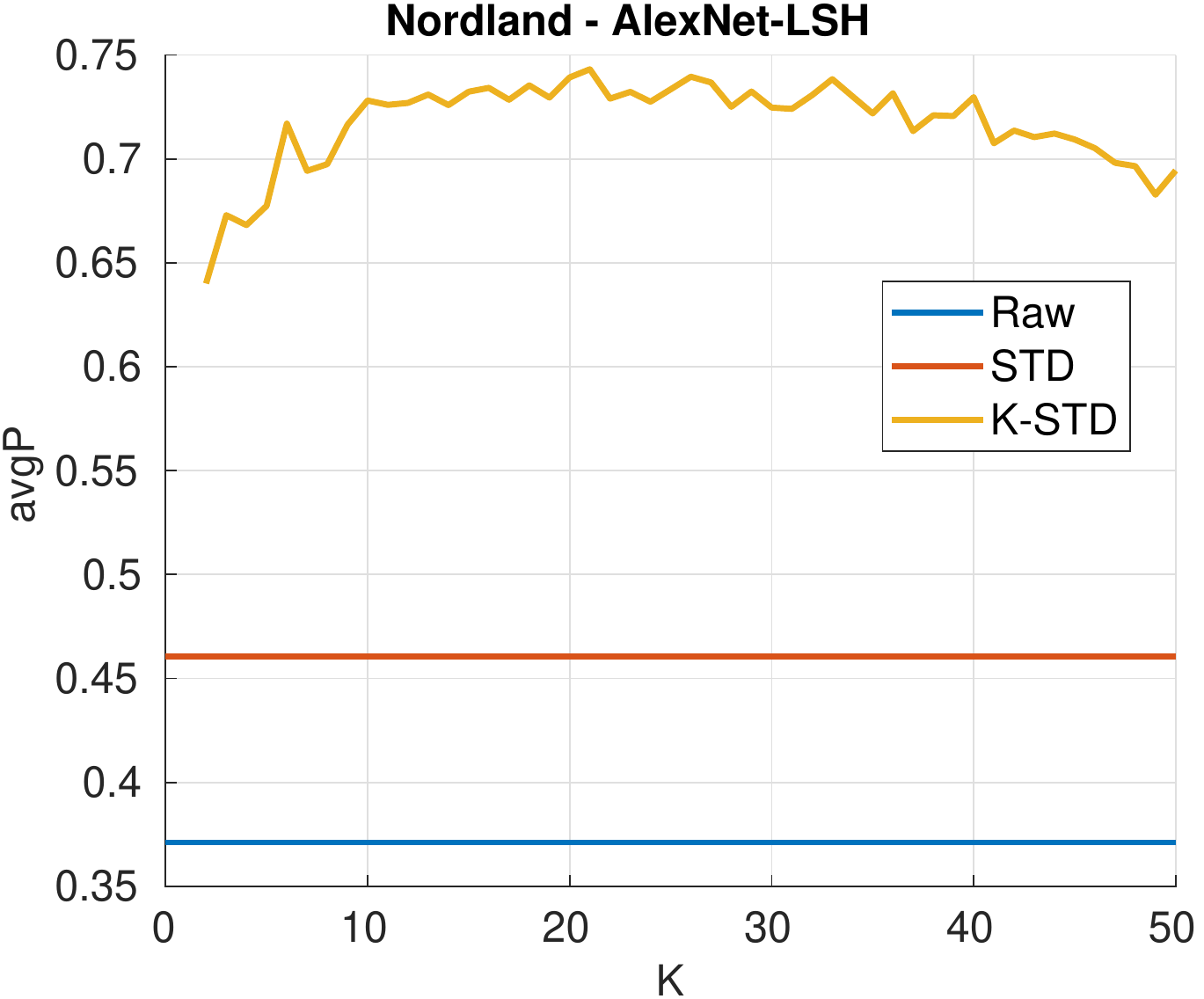}
 \includegraphics[width=0.45\linewidth]{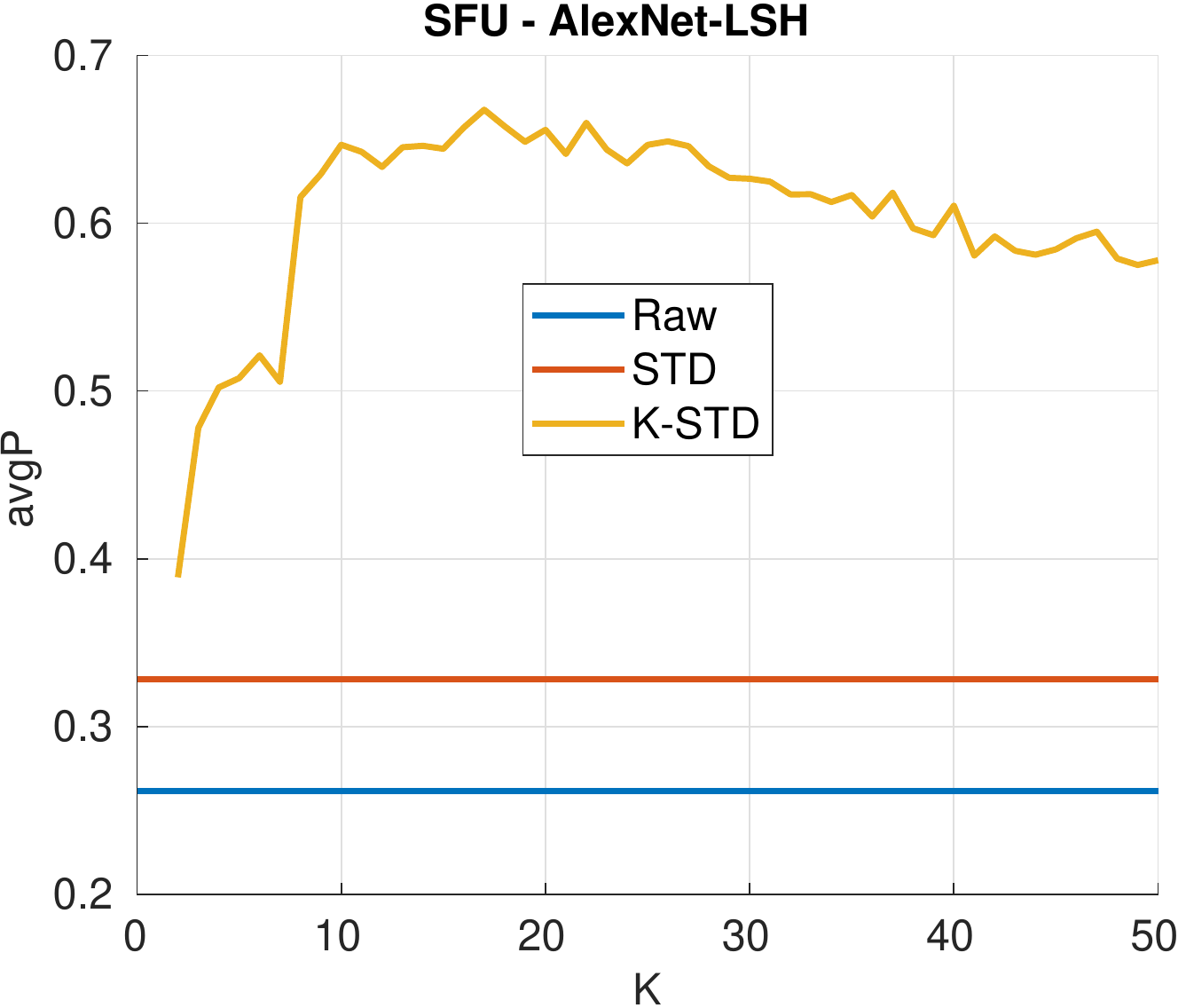}
 \includegraphics[width=0.45\linewidth]{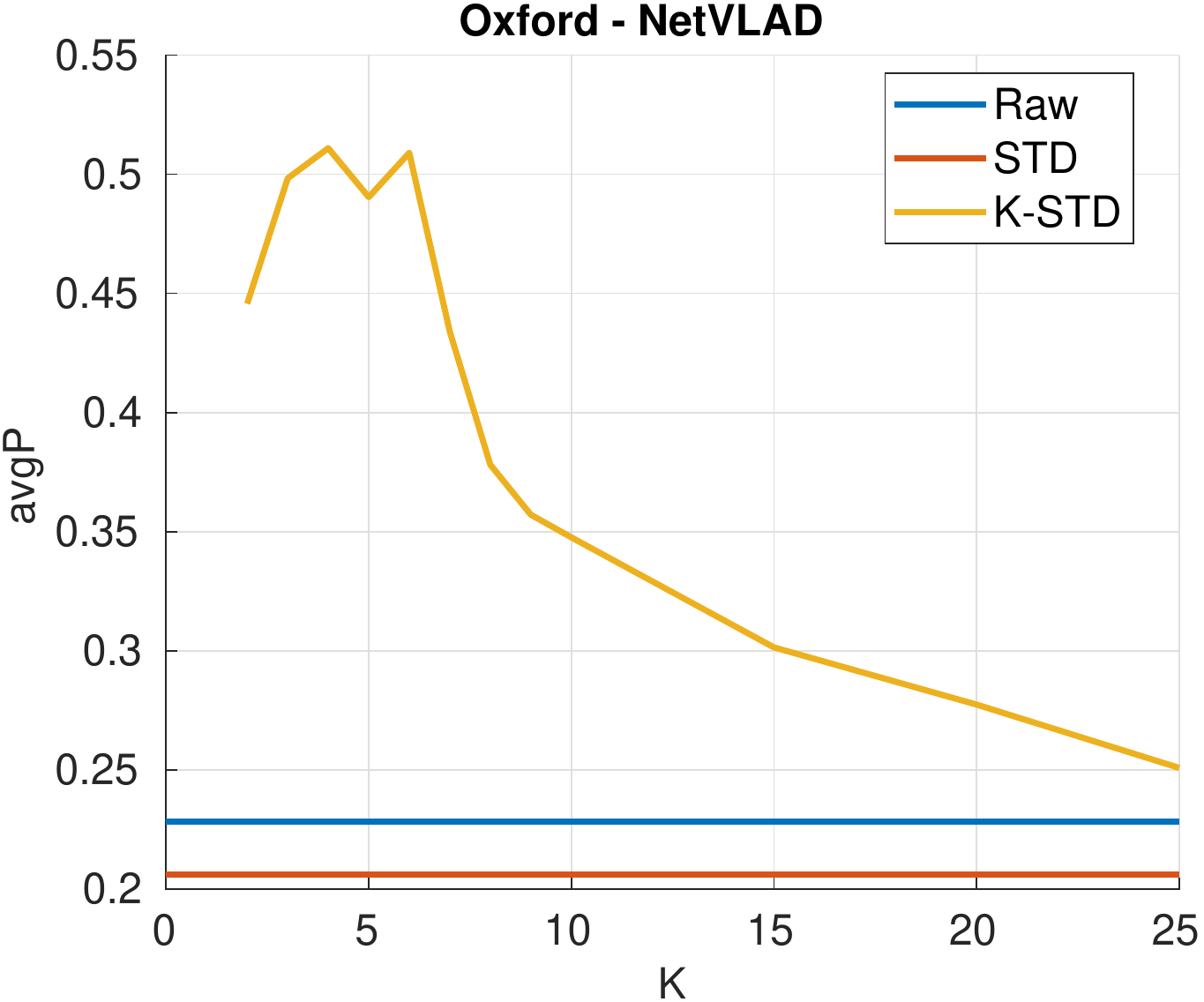}
 
 \vspace{-0.3cm}
 
 \caption{Evaluation of the average precision over $K$ for the clustering based method K-STD on the multi-conditions dataset with continuous condition changes. Note that only the better performing CNN-descriptor is shown for each dataset; the curves for the other descriptors look similarly with lower performance.}
\vspace{-0.5cm}
 \label{fig:k}
\end{figure}

As could be seen in the previous section, the application of standardization on $K$ clusters in the data can improve the performance remarkably in comparison to the raw and standardized descriptors.
Fig.~\ref{fig:k} shows the performance of K-STD over its parameter $K$ on the \textit{multi-conditions datasets} for the better performing CNN-descriptor.
While the performance for Nordland and SFU is relatively stable over a range of $K$, the range is lower for Oxford.
However, K-STD always achieves better performance than the raw and standardized descriptors over the evaluated range of $K$.

In our experiments, we used k-means with cosine distance for clustering, but better performing clustering algorithms could exist.
For the choice of $K$, we tried a parameter estimation via \textit{gap statistic} \cite{Tibshirani2001}, but it does not work for high-dimensional problems and returns always the maximum allowed $K$.

\subsection{Evaluation of parameters $q$ and $p$ for DR and CR}
\begin{figure}[tb]
 \centering

 \includegraphics[width=0.49\linewidth, trim=0.2cm 0 0.7cm 0, clip]{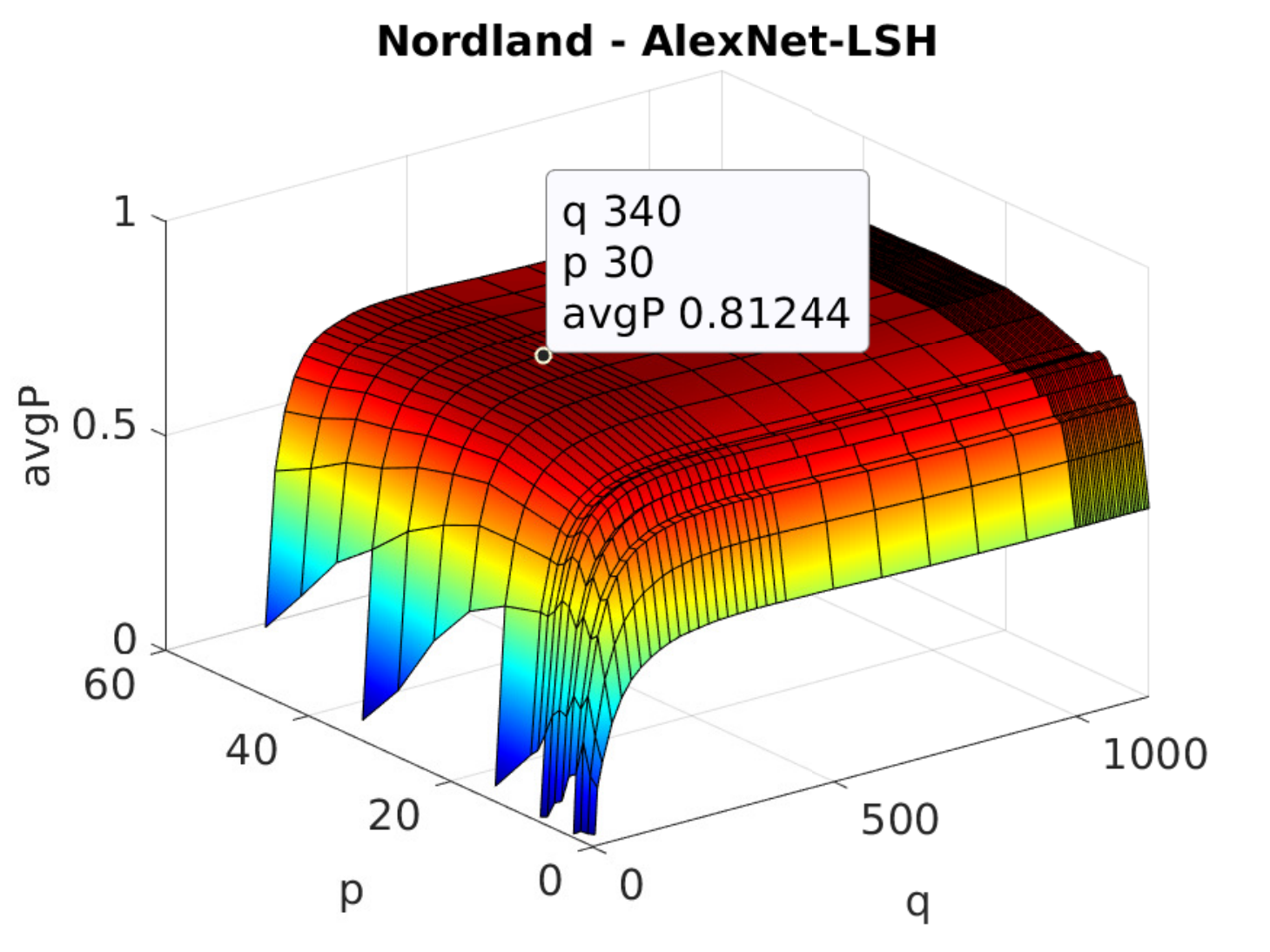}
 \includegraphics[width=0.49\linewidth, trim=0.2cm 0 0.7cm 0, clip]{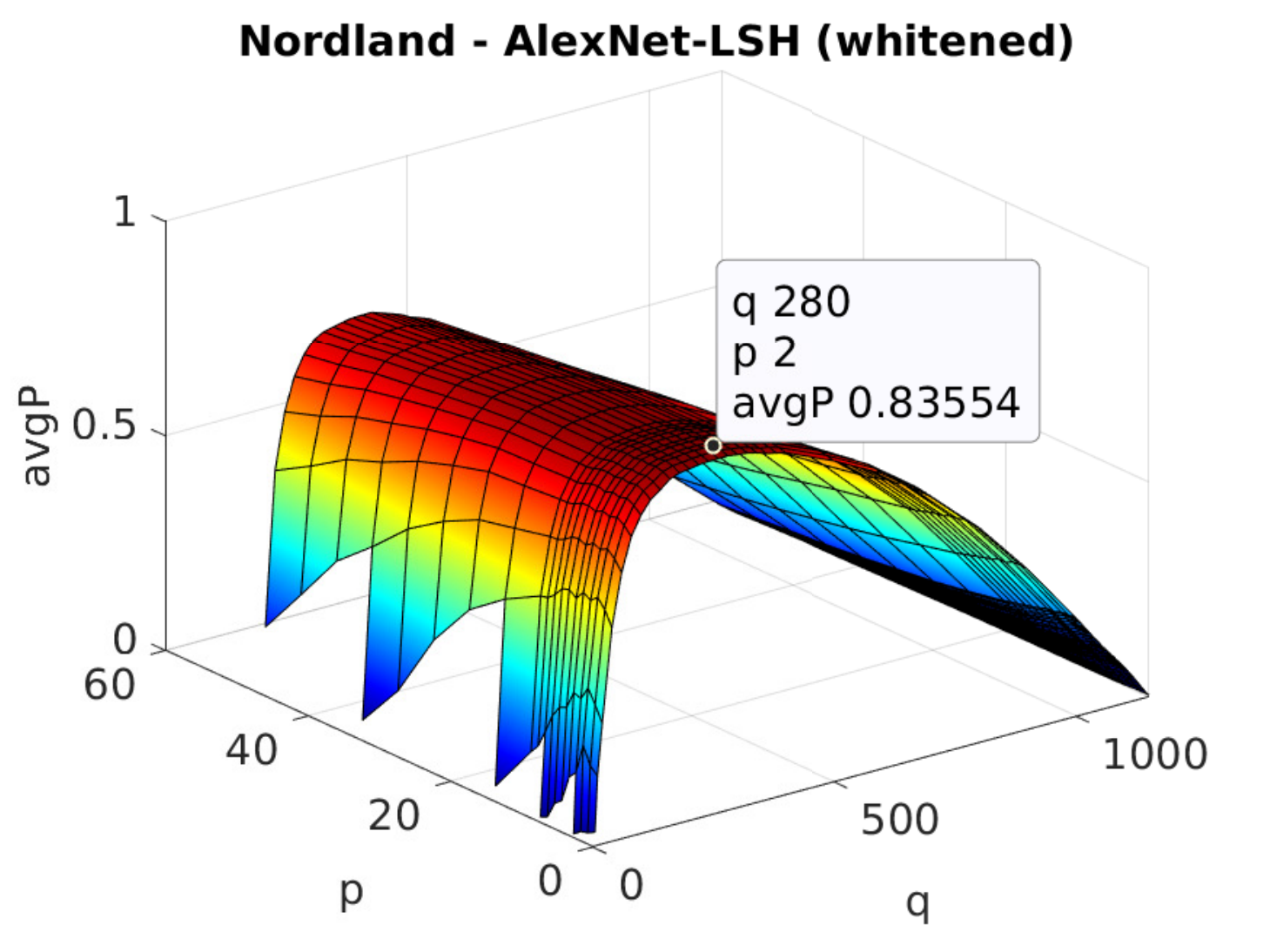}
 
 \includegraphics[width=0.49\linewidth, trim=0.2cm 0 0.7cm 0, clip]{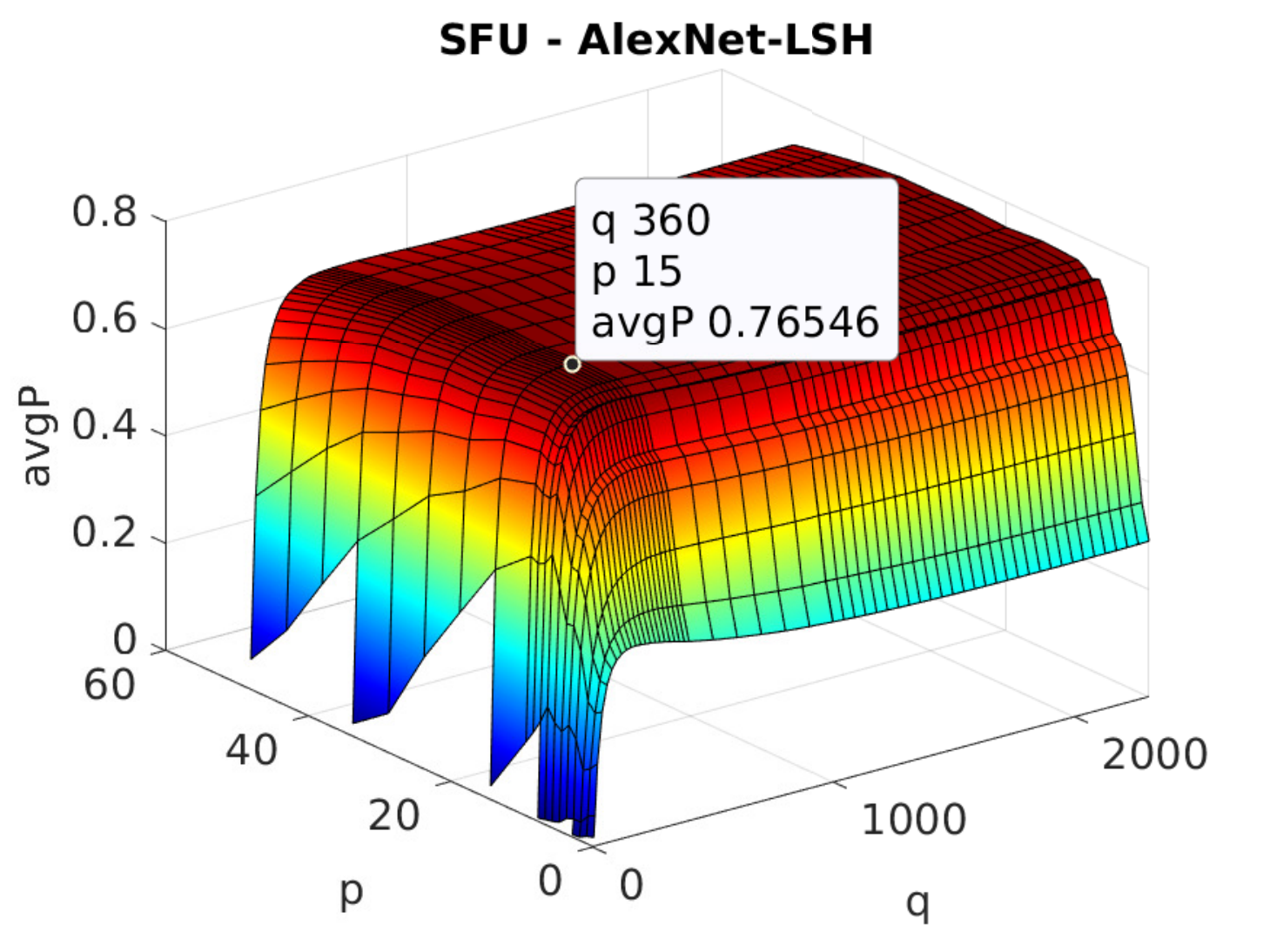}
 \includegraphics[width=0.49\linewidth, trim=0.2cm 0 0.7cm 0, clip]{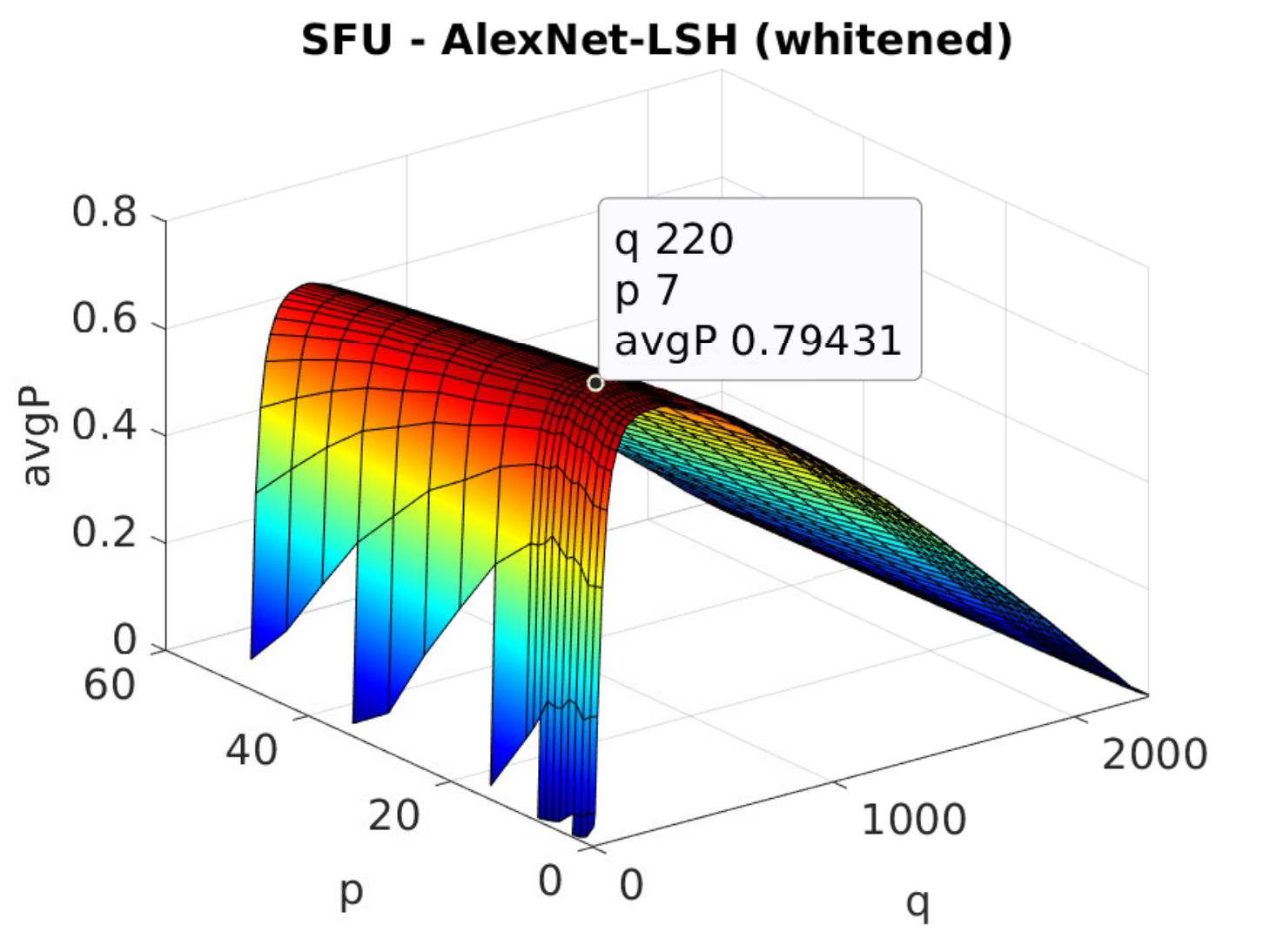}
 
 \includegraphics[width=0.49\linewidth, trim=0.2cm 0 0cm 0, clip]{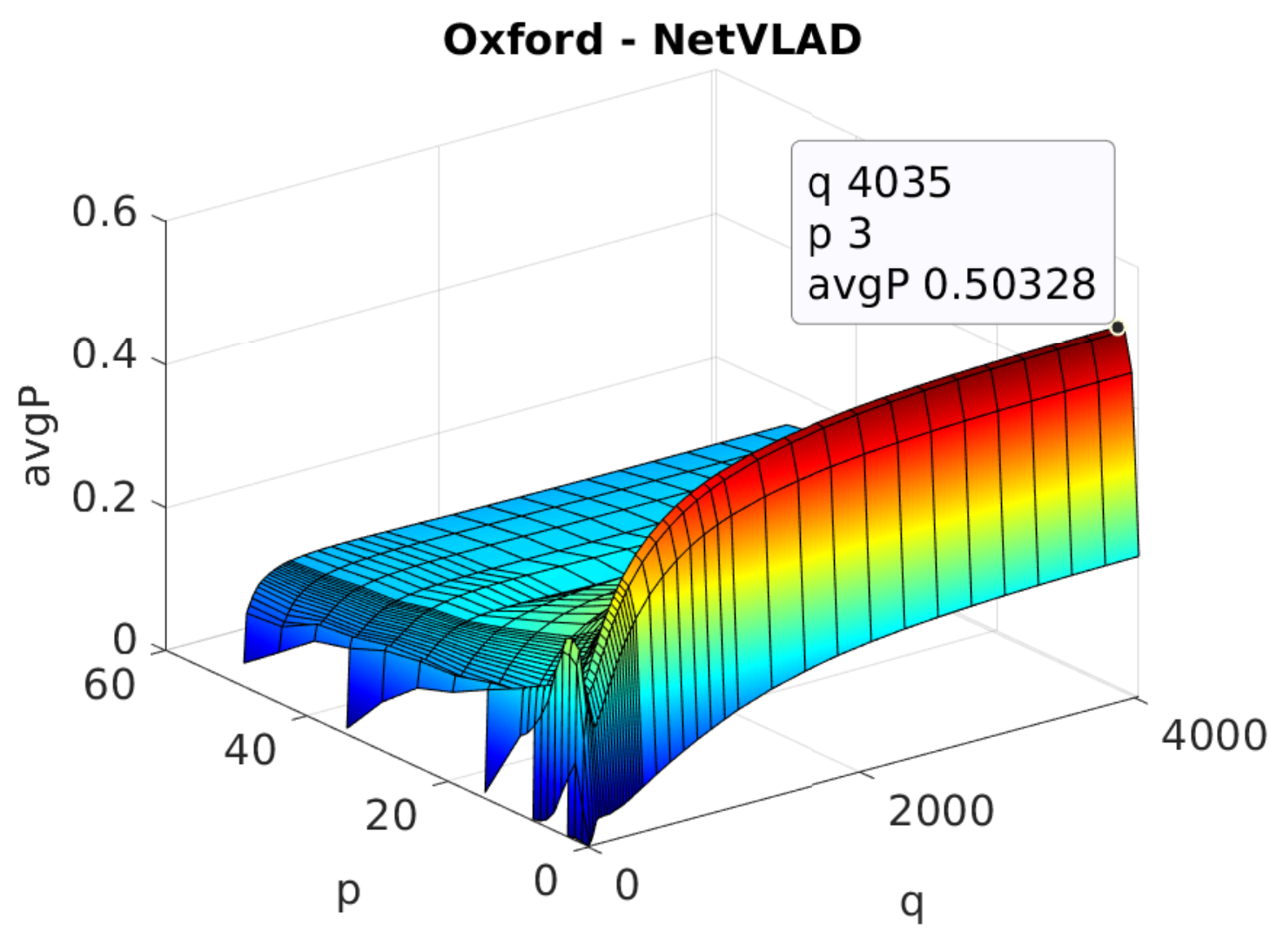}
 \includegraphics[width=0.49\linewidth, trim=0.1cm 0 0cm 0, clip]{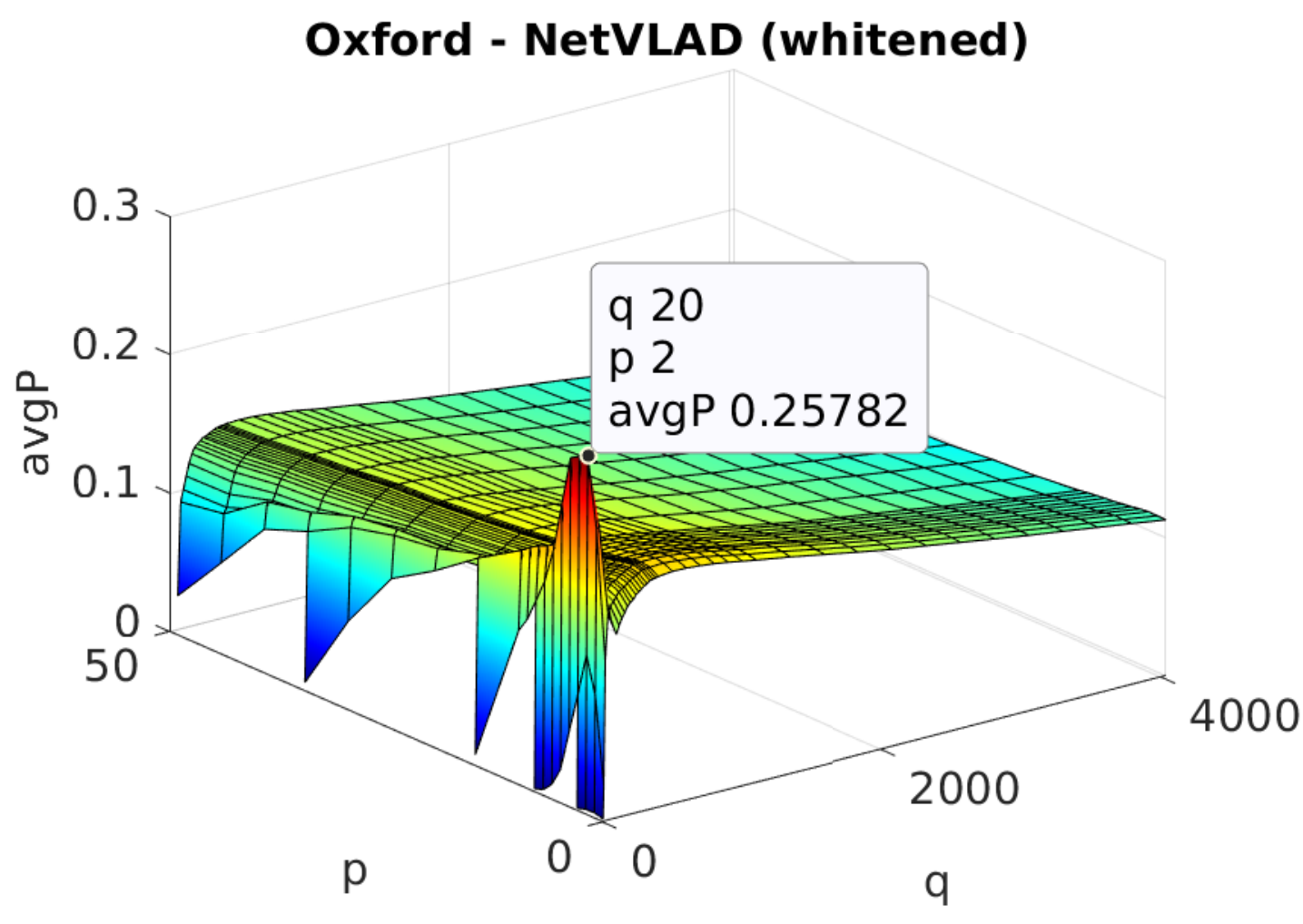}
 \vspace{-0.6cm} 
 \caption{Evaluation of the average precision over the parameters $q$ and $p$ for the PCA-based DR and CR on the multi-conditions datasets with continuous condition changes. The data points shows the maximum performance. Note that only the better performing CNN-descriptor is shown for each dataset; the surfaces for the other descriptors look similarly with lower performance.}
\vspace{-0.4	cm}
 \label{fig:pq_mixed}
\end{figure}

As shown in Sec.~\ref{sec:exp_all}, DR and CR can improve the performance remarkably in comparison to the raw and standardized descriptors.
Fig.~\ref{fig:pq_mixed} shows the performance of the combination of DR and CR over their parameters $q$ and $p$ on the \textit{multi-conditions datasets} for the better performing CNN-descriptor without and with whitening as proposed in \cite{Lowry2016}.
We used economic/reduced SVD to improve memory efficiency.
This is why the maximum $q$ is the minimum of either the number of descriptors or the number of features in a descriptor.
The results for $p=0$ are also shown.
Accordingly, the performance of stand-alone algorithms, DR or CR, can also be seen in the plots.

The range of good performing $q$ and $p$ is quite high without whitening for Nordland and SFU; in contrast, for Oxford there there is only a low range for $p$.
With whitening, the range of good $q$ is also limited but it can achieve slightly better results on Nordland and SFU, while it performs worse on Oxford.
\cite{Lowry2016} show a performance degradation for CR with whitening, but our results reveal that they would probably have to introduce DR in their approach.
To be memory and computationally efficient, DR could always be used; with whitening $q$ is even lower.

The different shapes of the performance curves over $q$ and $p$ between Nordland/SFU and Oxford show the dependency on the underlying data - there might be no generally optimal parameters for DR and CR.

\section{CONCLUSION}
In this paper, we discussed the problem of place recognition under in-sequence condition changes.
We defined three types of in-sequence condition changes in Sec.~\ref{sec:cond_change}, and described three approaches based on unsupervised learning that are suited for this problem in Sec.~\ref{sec:approaches}.
We could show in a broad range of datasets and sequence-combinations in Sec.\ref{sec:raw_descriptors} that raw CNN-descriptors perform well on datasets with only one condition in reference and query, and that standardization always maintains and often improves the results further.
However, as soon as in-sequence condition changes were introduced both approaches failed.
Experiments in Sec.~\ref{sec:exp_all} on three datasets with multiple condition changes showed that the three unsupervised learning based methods can remarkably improve the results as well as that the proposed novel approach K-STD creates state-of-the-art results.
Finally, we evaluated the performance of the three methods over their parameters $K$, $q$ and $p$.
The estimation of optimal values for these parameters is part of future work.
The performance of K-STD could potentially be improved with an alternative clustering approach.
Since all methods take as input descriptors and return more condition-invariant descriptors, they can be easily combined with available sequence-based methods (e.g., \textit{SeqSLAM} \cite{Milford2012}, \textit{HMM} \cite{Hansen2014}, \textit{ABLE-M} \cite{Arroyo2015}, \textit{MCN} \cite{Neubert2019}) or \textit{experience maps} \cite{Churchill2013}.
For future investigations on continuous condition changes, we are going to record datasets with changing conditions, e.g. while sunset and sunrise.


\end{document}